# TÜRKÇE İÇİN DERİN ÖĞRENME TABANLI DOĞAL DİL İŞLEME MODELİ GELİŞTİRİLMESİ

Barış BABÜROĞLU[a], Adem TEKEREK[b], Mehmet TEKEREK[c]

[a]*Kahramanmaraş Sütçüimam Üniversitesi, Enformatik Anabilim Dalı, 46040, Kahramanmaraş, Türkiye, barisbaburoglu@gmail.com*
[b]*Gazi Üniversitesi, Bilgi İşlem Dairesi Başkanlığı, 06560, Ankara, Türkiye, atekerek@gazi.edu.tr*
[c]*Kahramanmaraş Sütçüimam Üniversitesi, Eğitim Fakültesi, Bilgisayar ve Öğretim Teknolojileri Eğitimi Bölümü, 46040, Kahramanmaraş, Türkiye, tekerek@ksu.edu.tr*

**Özet—** Doğal dil, insanları diğer canlılardan ayıran ve insanların iletişim kurmasını sağlayan en temel özelliklerden biridir. Dil, insanın duygu ve düşüncelerini ifade etmede kullandığı ve kültürlerin nesiller boyunca aktarılmasını sağlayan bir araçtır. Günlük hayatta karşılaşılan yazılar ve sesler birer doğal dil örneğidir. Doğal dilde birçok kelime zamanla yok olurken diğer taraftan yeni kelimeler de türetilmektedir. Bu yüzden doğal dil işleme (DDİ) süreci insan için bile karmaşık yapıya sahipken, bilgisayar ortamında işlenmesi de zor olmaktadır. İnsanların dili nasıl kullandığını dil bilim alanı incelemektedir. Dil bilimciler ve bilgisayar bilimcilerinin ortak çalışmasını gerektiren doğal dil işleme çalışmaları, insan bilgisayar etkileşiminde önemli rol oynamaktadır. Doğal dil işleme çalışmaları, yapay zekâ teknolojilerinin, dil bilimi alanında kullanılması ile artmıştır. Yapay zekâ çalışma alanlarından olan derin öğrenme yöntemleri ile doğal dile yakın seviyede platformlar geliştirilmektedir. Dili anlama, makine çevirisi ve sözcük etiketleme için geliştirilen platformlar derin öğrenme yöntemlerinden faydalanmaktadır. Derin öğrenme mimarilerinden olan özyinelemeli sinir ağları (Recurrent Neural Network - RNN), metin veya ses verileri gibi sıralı verileri işlemede tercih edilmektedir. Bu çalışmada bir RNN türü olan iki yönlü uzun-kısa vadeli bellek (Bidirectional Long Short - Term Memory - BLSTM) kullanılarak Türkçe sözcük etiketleme modeli önerilmiştir. Önerilen sözcük etiketleme modeli, doğal dil araştırmacılarına, kendi analizlerini gerçekleştirme ve kullanabilme imkânı verecek bir platform ile sunulmaktadır. İki yönlü LSTM kullanılarak geliştirilen platformun geliştirilme aşamasında uzman görüşü ile geri bildirimler alınarak, sözcük etiketleyicinin hata oranı azaltılmıştır.

**Anahtar Kelimeler—** Doğal Dil İşleme, Derin Öğrenme, İki Yönlü Uzun-Kısa Vadeli Bellek, Özyinelemeli Sinir Ağları, Sözcük Etiketleme

# DEVELOPMENT OF DEEP LEARNING BASED NATURAL LANGUAGE PROCESSING MODEL FOR TURKISH

**Abstract—** Natural language is one of the most fundamental features that distinguish people from other living things and enable people to communicate each other. Language is a tool that enables people to express their feelings and thoughts and to transfers cultures through generations. Texts and audio are examples of natural language in daily life. In the natural language, many words disappear in time, on the other hand new words are derived. Therefore, while the process of natural language processing (NLP) is complex even for human, it is difficult to process in computer system. The area of linguistics examines how people use language. NLP, which requires the collaboration of linguists and computer scientists, plays an important role in human computer interaction. Studies in NLP have increased with the use of artificial intelligence technologies in the field of linguistics. With the deep learning methods which are one of the artificial intelligence study areas, platforms close to natural language are being developed. Developed platforms for language comprehension, machine translation and part of speech (POS) tagging benefit from deep learning methods. Recurrent Neural Network (RNN), one of the deep learning architectures, is preferred for processing sequential data such as text or audio data. In this study, Turkish POS tagging model has been proposed by using Bidirectional Long-Short Term Memory (BLSTM) which is an RNN type. The proposed POS tagging model is provided to natural language researchers with a platform that allows them to perform and use their own analysis. In the development phase of the platform developed by using BLSTM, the error rate of the POS tagger has been reduced by taking feedback with expert opinion.

## 1. GİRİŞ

Doğal dil, insanların kendilerini ifade etmeleri ve iletişim kurabilmeleri için kullanılan bir araçtır. Chomsky, dilin çocukluk yıllarında duyulandan, doğal bir dile dönüşümünün, insanın genetik yapısıyla ilişkili olduğunu ifade etmektedir (Chomsky, 1986). İnsanların dili nasıl edindiği, ürettiği ve anladığı dil biliminin araştırma alanıdır. Dil bilimciler dilsel ifadeleri işlemek için kural tabanlı yaklaşımlar öne sürmüşlerdir. Ancak dil kullanımının üreticiliği kurallara her zaman uymamaktadır. Bu doğrultuda dil ifadelerine dilbilgisi kuralları uygulamak yerine istatistiksel yaklaşımlar uygulanarak dil kullanımında ortak kalıplar elde edilmeye çalışılmıştır. Dilin istatistiksel modelleri, dil bilimi ve bilgisayar biliminin alt bilim dalı olan doğal dil işleme (DDİ) çalışmalarında başarı ile uygulanmıştır (Schütze & Manning, 1999). DDİ'nin amacı, doğal dilleri otomatik olarak oluşturma ve anlamadaki problemleri incelemektir (Young, Hazarika, Poria, & Cambria, 2018). DDİ insanlar tarafından üretilen sesleri ve metinleri işleyerek insan bilgisayar etkileşiminin sağlanmasına yardımcı olmaktadır.

DDİ insan dilinin otomatik analizi ve gösterimi için teorik olarak motive edilmiş hesaplama teknikleridir (Cambria & White, 2014). Her yaşta insanın sosyal medyaya ulaşabildiği bir ortamda, üretilen veri miktarı, her geçen gün artarak devam etmektedir. İnsanlar tarafından doğal olarak oluşturulan veriler, doğrudan işlenecek durumda değildir. Bu yüzden insan makine iletişimini sağlamak için verileri anlamlandırma ve verimli kullanabilme çabası birçok alanın birlikte çalışmasını gerektirmiştir. İlk zamanlarda yapılan çoğu DDİ çalışmaları, tek tek sözcüklere odaklanmışken 19. yüzyılın sonlarına doğru sözcüklerin birbirleriyle olan ilişkisine ve bütün üzerinde anlam bilim çalışmalarına yönelmiştir (Cambria & White, 2014).

DDİ çalışmaları ilk olarak metni anlamak için ses veya metinden özellik çıkarımı yapan bir ön işlemeden geçmektedir. Ardından şekilbilim (morphological), sözdizim (syntactic), anlambilim (semantic) ve söylev (discourse) işleme çalışmaları gerçekleştirilebilmektedir. Bu çalışma alanları sözcük kökleri, sözcük bağlamları ve anlambilim açısından bazı zorluklara sahiptir. Zorlukları aşmak için geliştirilen dilbilgisine dayalı kural tabanlı DDİ çalışmaları (Brill 1992), (J. Gimenez and L. Marquez 2004), el yapımı özelliklere dayanmaktadır. El yapımı özellikler zaman almakta ve yetersiz kalmaktadır (Young, Hazarika, Poria, & Cambria, 2018). Eksikliklerin giderilmesi için geliştirilen yapay zekâ yöntemlerinin önemli bir uygulaması olan derin öğrenme, yapay sinir ağı yapısı, güçlü donanımı ve büyük veri girdisi ile daha iyi sonuçlar elde edilmesini sağlamıştır (Song & Lee, 2013).

1965 yılında derin öğrenmenin temeli kabul edilen çok katmanlı bir perceptron türü algoritma (Ivakhnenko & Lapa, 1965) önerilmiştir. Fakat o yıllardaki basit bir ağın eğitiminin uzun sürmesi ve yüksek hesaplama maliyetlerinden dolayı, destek vektör makinaları gibi el ile hazırlanmış özelliklere sahip modeller (Cortes & Vapnik, 1995)) kabul görmüştür (Şeker, Dirib, & Balık, 2017). Yakın zamanda grafik işleme birimi (GPU) ve diğer donanımsal gelişmeler sayesinde hesaplama maliyetleri düştüğünden, çok sayıda gizli katmandan oluşan yapay sinir ağları tekrar kullanılmaya başlanmıştır (Schmidhuber, 2015). Bu doğrultuda DDİ çalışmalarında, makine çevirisi, bilgi alma, metin özetleme, soru cevaplama, bilgi çıkarma, konu modelleme ve sözcük etiketleme gibi görevlere, derin öğrenme uygulanmasına odaklanılmaktadır (Young, et al., 2018) (Şeker, Dirib, & Balık, 2017).

Basit bir derin öğrenme çerçevesinin, adlandırılmış varlık tanıma, anlamsal rol etiketleme ve sözcük etiketleme gibi birçok DDİ görevinde, en modern yaklaşımlardan daha iyi performans gösterdiği ortaya konulmuştur (Collobert, ve diğerleri, 2011). DDİ alanında istatistiksel yöntemler, kural tabanlı yöntemlerden daha başarılı olmaktadır. Bu alanda sözcük etiketlemesi, sözcük türlerinin sözcüklere atanması ile gerçekleştirilmektedir (sözcük/isim, sıfat, fiil, vb.). Etiketler bilgisayarların cümlede ifade edileni anlamasında kolaylık sağlamaktadır. Ancak sözcükler farklı bağlamlarda kullanıldığı zaman farklı anlamlar ifade edebilmektedir. Örneğin 'yüz' sözcüğü kullanıldığı bağlama göre isim veya fiil etiketini alabilmektedir. Verilen örneği incelediğimizde, her sözcüğü etiketlemenin söz konusu olmadığı görülmektedir. Sözcük etiketlemede belirsizliği gidermek için etiketlenecek sözcüğün öncesinde kullanılan sözcüklere (geri yönde) ve sonrasında kullanılan sözcüklere(ileri yönde) bakılarak sözcük, doğru etiket sınıfına eklenebilmektedir. Etiketleme işlemlerinde, iki yönde ki bilgileri kullanarak olasılık üreten, bir tekrarlayan sinir ağı türü olan, iki yönlü uzun-kısa vadeli bellek (BLSTM) ağlarının, sıralı verileri etiketlemek için doğal dil işleme çalışmalarında çok etkili olduğu gösterilmektedir (Wang, Qian, K. Soong, He, & Zhao, 2015). BLSTM mimarisi, dil modelleme (Sundermeyer, Schlüter, & Ney, 2012), (Sundermeyer, Ney, & Schluter, 2015), dili anlama (Yao, Zweig, Hwang, Shi, & Yu, 2013), makine çevirisi (Sundermeyer, Alkhouli, Wuebker, & Ney, 2014) ve sözcük etiketlemesi gibi doğal dil işleme alanındaki uygulamalar için üstün performans elde edilmesine yardımcı olmaktadır (Wang, Qian, K. Soong, He, & Zhao, 2015).

Bu çalışma 4 bölümden oluşmaktadır. 2. bölümde çalışmada kullanılan derin öğrenme yöntemleri anlatılmıştır. 3. bölümde önerilen modelin geliştirilme süreçlerinden bahsedilmiştir. 4. bölümde ise sonuçlara değinilmiştir.

## 2. DERİN ÖĞRENME

Derin öğrenme, özellik çıkarma için ardışık işlem birimi katmanları kullanan ve her katman çıktısı bir sonraki katmanın girişini besleyen bir yapay sinir ağı türüdür (Deng & Yu, 2014). Yüksek hesaplama ve veri işlemeye ihtiyaç duyan alanlarda başarılı sonuçlar sunan derin öğrenme yöntemleri, DDİ araştırmalarında da başarı göstermektedir. DDİ problemlerini hedef alan makine öğrenme yaklaşımları, çok yüksek boyutlu ve seyrek özellikler üzerine eğitilmiş sığ modellere (örneğin, destek vektör makinaları ve lojistik regresyon) dayanmaktadır. Ancak vektör temsillerine dayanan derin sinir ağları, çeşitli DDİ problemlerinde başarılı sonuçlar göstermektedir. Bu sonuçlar, sözcük ekleme ve derin öğrenme yöntemlerinin başarısı ile ortaya çıkmıştır (Mikolov, Karafiat, Burget, Cernocky, & Khudanpur, 2010) (Mikolov, Sutskever, Chen, Corrado, & Dean, 2013).

Derin öğrenme yöntemlerinin, DDİ problemlerinde iyi performans göstermesi ile zor DDİ çalışmalarını çözmek için Konvolüsyonel sinir ağları (CNNs), tekrarlayan sinir ağları ve özyinelemeli sinir ağları(RNNs) gibi temel derin öğrenme ile ilgili modeller önerilmektedir (Young, Hazarika, Poria, & Cambria, 2018). Peilu Wang ve arkadaşlarının yaptığı bir çalışmada (2015), sözcük etiketleme görevi için sözcük ekleme ile Çift Yönlü Uzun Kısa Vadeli Bellek Tekrarlayan Sinir Ağı (BLSTM-RNN) kullanımı önerilmiştir. BLSTM-RNN'in konuşma ifadeleri, el yazıları veya metin verileri gibi sıralı verileri etiketlemek için çok etkili olduğu gösterilmiştir.

### 2.1. Uzun Kısa Vadeli Bellek Ağları (LSTMs)

LSTM'ler uzun vadeli bağımlılık problemini çözmek için tasarlanan özel bir RNN türüdür (Hochreiter & Schmidhuber, 1997). Bilgiyi uzun süre hatırlama davranışı sergilerler. LSTM'ler tüm tekrarlayan sinir ağları gibi bir sinir ağının tekrar eden hücrelerine sahiptir, fakat tekrar eden hücre tek bir sinir ağı kapısına sahip olmak yerine, etkileşime giren 3 adet kapıya (Şekil 1) sahiptir (Graves & Schmidhuberab, 2005).

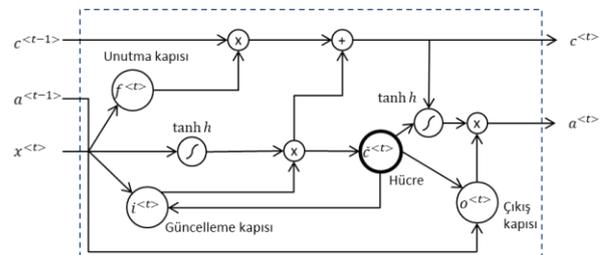

Şekil 1. Bir Uzun-Kısa Vadeli Hafıza Hücresi

Bir LSTM davranışını tanımlayan kapılar güncelleme (update), unutma (forget) ve çıkış (output) kapıları olarak adlandırılmaktadır. Burada $i^{<t>}$, $f^{<t>}$ ve $o^{<t>}$ sırasıyla t anındaki güncelleme, unutma ve çıkış kapılarını, $c^{<t>}$, hücre durumunu ve a<sup>\<t\></sup>, tüm işlemler sonunda karar verilen bilgiyi belirtir.

Bir sigmoid(σ) işlevine sahip unutma kapısı (eşitlik 1) giriş ($x^{<t>}$) ve önceki hücre durumu ($a^{<t-1>}$) bilgisine bakarak bilginin unutulup unutulmayacağına karar verir.

$$f^{<t>} = \sigma(W_f[a^{<t-1>}, x^{<t>}]) + b_f \quad (1)$$

Bir sigmoid(σ) işlevine sahip güncelleme kapısı (eşitlik 2) hangi bilgilerin güncelleneceğine karar verir.

$$i^{<t>} = \sigma(W_u[a^{<t-1>}, x^{<t>}]) + b_u \quad (2)$$

Bir $\tanh$ işlevi, hücre durumuna ($c^{<t>}$) eklenebilecek yeni hücre durumu aday bilgileri ($\tilde{c}^{<t>}$) (eşitlik 3) vektörünü oluşturur.

$$\tilde{c}^{<t>} = \tanh(W_c[a^{<t-1>}, x^{<t>}]) + b_c \quad (3)$$

Dolayısıyla, mevcut hücre durumu ($c^{<t>}$) (eşitlik 4), hem önceki hücre durumu ($\tilde{c}^{<t>}$) hem de hücre tarafından üretilen mevcut bilgileri kullanılarak elde edilmiş olur.

$$c^{<t>} = i^{<t>} * \tilde{c}^{<t>} + f^{<t>} * c^{<t>} \quad (4)$$

Bir sigmoid(σ) işlevine sahip çıkış kapısı (eşitlik 5) hücre durumunun hangi kısımlarını üreteceğimize karar verir. Daha sonra hücre durumu ($c^{<t>}$) $\tanh$ işlevi ile filtrelenir ve çıkış kapısının çıktısı ile çarpılır. Bu durumda sadece karar verilen bilgiler ($a^{<t>}$) (eşitlik 6) elde edilir.

$$o^{<t>} = \sigma(W_o[a^{<t-1>}, x^{<t>}]) + b_o \quad (5)$$

$$a^{<t>} = o^{<t>} \tanh(c^{<t>}) \quad (6)$$

Standart LSTM ağları dizileri geçici sırayla işler, gelecekteki bağlamı görmezden gelirler. (Schuster & K. Paliwal, 1997)'da önerilen iki yönlü RNN gizli bağlantıların ileriye doğru sırayla aktığı, ikinci bir katman uygulayarak tek yönlü LSTM ağlarını genişletmektedir.

## 2.2. İki Yönlü Uzun Kısa Vadeli Bellek Ağları (BLSTMs)

İki Yönlü Özyinelemeli Sinir Ağları (BRNNs) bir dizinin hem önceki zamanlardaki hem de ileriki zamanlardaki bilgilerini kullanabilmeyi sağlar. Bu yöntem, normal bir RNN'nin durum nöronlarını pozitif zaman yönünden (ileri durumlar- $\vec{a}$) ve negatif zaman yönünden (geri durumlar- $\overleftarrow{a}$) birbiriyle bağlantısı olmayan iki duruma (şekil 2) bölmektedir (Schuster & K. Paliwal, 1997).

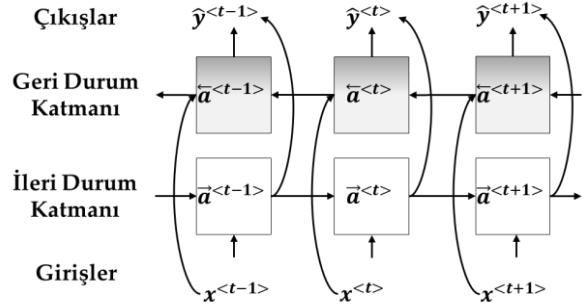

Şekil 2. İki Yönlü Özyinelemeli Sinir Ağının (BRNN) 3 Zaman Adımı

Böylece normal bir RNN'den farklı olarak iki katmandan gelen değerler ile hesaplama yapılır. Geri durumlar ve ileri durumlardan elde edilen değerler, ağırlıklar ve koşullu olasılık bayes değerlerinin hesaplanmasıyla t anındaki $\hat{y}^{<t>}$ değeri (eşitlik 7) elde edilmiş olur.

$$\hat{y}^{<t>} = g(W_y[\vec{a}^{<t>}, \overleftarrow{a}^{<t>}]) + b_y \quad (7)$$

İki yönlü özyinelemeli sinir ağlarında tekrar eden hücrelerde LSTM hücresi kullanıldığında İki yönlü uzun-kısa vadeli bellek ağları (BLSTMs) mimarisi (Şekil 3) elde edilmektedir (Graves & Schmidhuberab, 2005).

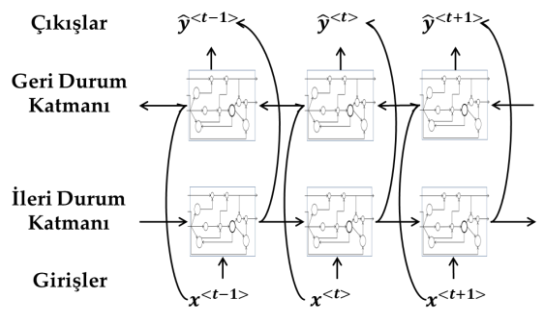

Şekil 3. İki Yönlü Uzun-Kısa Vadeli Bellek Ağının ileri ve geri yönde zaman dizisi

Bu mimaride ileri ve geri yöndeki hesaplamalar aynı anda çalıştırılır ve iki yönde yapılan hesaplamalar sonucu ulaşılan bilgiler birleştirilerek çıktı sonucuna ulaşılır. Bu şekilde iki yöndeki bilgilerin kullanılması sıralı verilerin işlenmesinde avantaj sağlar. (Graves & Schmidhuberab, 2005)'deki araştırmada iki yönlü ağların tek yönlü ağlardan daha etkili olduğunu gösterilmiştir. İki yönlü uzun-kısa vadeli bellek(BLSTM) ağları, doğal dil işleme çalışmalarında da başarılı şekilde kullanılmaktadır (Wang, Qian, K. Soong, He, & Zhao, 2015).

Günümüzde DDİ alanında çalışan araştırmacılar, derin öğrenme yöntemlerinin kullanımına yönelmektedir. Derin öğrenme yöntemleri kullanılarak geliştirilen doğal dil işleme çalışmalarının İngilizce üzerine yoğunlaştığı görülmektedir (Şeker, Dirib, & Balık, 2017). Türkçe için derin öğrenme yöntemleri ile doğal dil işleme çalışmaları sınırlıdır. Türkçe sözcük etiketleme için derin öğrenme ile eğitilmiş bir model henüz bulunmamaktadır. Bu çalışmada, RNN mimarisinde iki yönlü uzun-kısa vadeli bellek kullanılarak Türkçe sözcük etiketleme için bir DDİ modelinin geliştirilmesi amaçlanmıştır. Platform ile dil bilim bilimi ve bilgisayar bilimi altında çalışan araştırmacılar için kaynak eksikliğinin giderilmesi hedeflenmektedir. Ayrıca platformun, insana özgü doğal dile yakın seviyede işleme gerçekleştirmeyi destekleyeceği düşünülmektedir.

## 3. ÖNERİLEN DDİ MODELİ

Bu çalışmada Keras ve TensorFlow ve iki yönlü LSTM RNN(BLSTM-RNN) derin öğrenme modeli kullanılmıştır. Yazılım geliştirilmesi için Python programlama dili kullanılmıştır.

BLSTM-RNN ile eğitimi gerçekleştirilen derlemin içeriğinde 35 konu ile işaretlenmiş, konu başına 200 belgeden oluşan, toplam 7000 adet belge bulunmaktadır. Her belgenin başlığı, özeti, (varsa) anahtar kelimeleri, kaynak ismi bulunmaktadır. (Ozturk, Sankur, Gungor, & Yilmaz, 2014)'deki çalışma ile hazırlanan bu belgelerin özet kısımları kullanılmıştır.

Keras kütüphanesinin kelimelerle veya etiketlerle değil sayılarla çalışması gerekmektedir. Her bir kelimeye ve etikete benzersiz bir tamsayı atanarak bir sözlükte dizine alınmıştır. Bu sözlükler kelime dağarcığı ve etiket dağarcığı olarak adlandırılabilir. Ayrıca bilinmeyen kelimeler için bir değer (OOV - Out Of Vocabulary) eklenmiştir. Keras yalnızca sabit boyutlu dizilerle ilgilendiği için veri setindeki en uzun cümle hesaplanmıştır. Buna göre diğer cümlelerde aynı boyutu sağlamak için bir değer (indeks olarak "0" ve karşılık gelen etiket olarak "-PAD-") eklenmiştir. Son olarak eğitim aşamasına geçilmeden önce Etiket dizileri One-Hot Encoded etiketlerinin dizilerine dönüştürülmüştür.

Toplanan özet kısımlar, standart bir trigram Saklı Markof Model konuşma parçacığı etiketleyici (HMM POS Tagger) ile eğitim verisinin sözcük etiketlemesi gerçekleştirilmiştir (Çöltekin, 2014). Elde edilen veriler NLTK Kütüphanesi kullanılarak büyük harf, sayılar ve noktalama işaretlerinden temizlenerek eğitime hazır forma dönüştürülmüştür.

Web tabanlı olarak geliştirilen Türkçe için iki yönlü LSTM-RNN ile doğal dil işleme platformunun mimarisi ve çalışma şekli sunulmaktadır (Şekil 4).

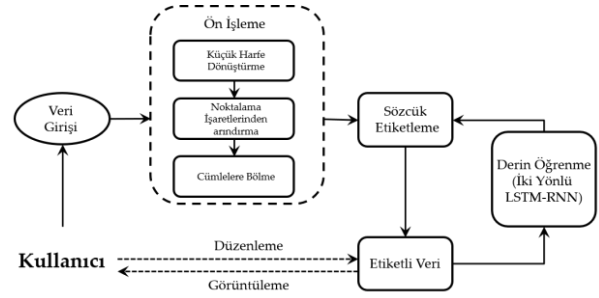

Şekil 4. İki Yönlü LSTM-RNN Türkçe Doğal Dil İşleme Platformu

Platform, araştırmacılara kaynak sağlaması ve kendi kendini geliştirmesi için kullanıcılara metin girdisi ve belge girdisi PDF olmak üzere 2 ayrı ekrandan veri girişi sağlamaktadır. Metin analizi ekranı, kullanıcılara el ile metin girme ve analiz gerçekleştirme imkânı sağlamaktadır. Belge analizi ekranı ile kullanıcılar PDF formatında belgeler yükleyerek verilerin analizlerini gerçekleştirebilmektedir. Bu iki ekrandan alınan veriler ilk olarak ön işleme sürecinden geçmektedir.

Ön işleme aşamasında yazılı belgelerden elde edilen verilerin işlenebilir hale getirilmesi için DDİ normalleştirme süreci uygulanır. Normalleştirme süreci küçük harfe dönüştürme, noktalama işaretlerinden arındırma ve sözcük parçalama (tokenize) işlemleridir. Veri ilk olarak küçük harfe dönüştürülür. İkinci aşamada, sözcük parçalama ile veriden cümleler elde edilir. Son olarak da veri, cümle içerisinde noktalama işaretlerinden arındırılarak sözcük etiketleme için hazır hale getirilir.

Ön işlemeden geçen veri, sözcük etiketleme aşamasında, iki yönlü LSTM ile analiz edilir. Analiz sonucu her sözcüğe cümle içerisindeki bağlamına göre en olası etiket atanır.

Etiketli veri kısmında, etiketi atanan sözcükler ve frekansları kullanıcıya sunulmaktadır.

Gerçekleştirilen analiz sonuçlarının dökümü alınarak Türkçe doğal dil işleme çalışmalarında kullanılabilir. Ayrıca kullanıcı düzenleme önerme seçeneği ile yanlış olduğunu düşündüğü etiketleri düzelterek platformun geliştirilmesine katkıda bulunabilir. Düzenleme yapılan etiketlenmiş veriler daha sonra eğitim aşamasına dâhil edilip etiketleme doğruluğunun arttırılması için kullanılır.

Şekil 5. Metin Analizi Ekranı

**4. SONUÇ**

Bu çalışmada, derin öğrenme yöntemleri kullanılarak dinamik ve kullanımı kolay bir web tabanlı DDİ modeli sunulmuştur. Herhangi bir el yapımı özellik veya dış araç kullanmadan sözcüklerin bağlamsal özelliklerinden faydalanarak etiket tahminleri gerçekleştirilmiştir. Bağlama bakılarak eğitim verisinde bulunmayan sözcükler tahmin edilmiştir. Tahmin için iki yönlü uzun-kısa vadeli bellek derin öğrenme mimarisi kullanılmıştır. Kullanıcılara tahmin edilen etiketleri düzeltme imkânı verilerek sistemin geliştirilmesine dâhil olmaları sağlanmıştır. Derin öğrenme kullanarak doğal dil işlemeyi gerçekleştiren ve kullanıcıları da sürece dâhil ederek doğal kavramının sürekliliğinin sağlayan Türk dilinde ilk platform olduğu düşünülmektedir. Geliştirilen modelde kullanılan eğitim verisinin miktarı arttıkça daha başarılı sonuçlar elde edileceğine inanılmaktadır.